\documentclass{article}

\usepackage[final,dblblindworkshop]{neurips_2026}

\workshoptitle{AI for Drug Discovery: Bridging the Translation Gap}

\usepackage[utf8]{inputenc}
\usepackage[T1]{fontenc}
\usepackage[draft]{hyperref}
\usepackage{url}
\usepackage{booktabs}
\usepackage{placeins} 
\usepackage{float}    
\usepackage{amsmath}
\usepackage{amssymb}
\usepackage{graphicx}
\usepackage{xcolor}



\title{Did the Model Forecast It, or Already Know It?
Evaluating Large Language Models for Novel Drug-Target
Prioritisation with a Temporal Knowledge-Graph Oracle}

\newcommand{\authline}[1]{{\small #1}}
\author{%
  Pui Chung Siu\thanks{Corresponding authors: Claudia Cabrera, Mani Mudaliar and
    Arkaitz Zubiaga.}\\\authline{Queen Mary University of London}\\\authline{Recursion Pharmaceuticals}\\\authline{\texttt{pui.siu@qmul.ac.uk}}
  \And
  Yuan Liang\\\authline{Queen Mary University of London}\\\authline{Recursion Pharmaceuticals}\\\authline{\texttt{yuan.liang@qmul.ac.uk}}
  \And
  Claudia Cabrera\\\authline{Queen Mary University of London}\\\authline{\texttt{c.cabrera@qmul.ac.uk}}
  \AND
  Mani Mudaliar\\\authline{Recursion Pharmaceuticals}\\\authline{\texttt{mani.mudaliar@recursion.com}}
  \And
  Arkaitz Zubiaga\\\authline{Queen Mary University of London}\\\authline{\texttt{a.zubiaga@qmul.ac.uk}}%
}

\begin{document}

\setlength{\tabcolsep}{2pt}
\maketitle
\setlength{\tabcolsep}{6pt}

\begin{abstract}
AI systems increasingly help decide which drug targets enter costly clinical
trials. We recast each target--disease hypothesis as a dated event, the pair's
first clinical trial. This lets us test whether a language model can forecast
such events after its training cutoff, rather than recall events before it. We
rebuild a dated biomedical knowledge graph as it stood at the end of each year
from 2020 up to its newest version. We then ask six open language models, with and without this
graph, which targets will enter a first trial for a disease the next year.
Without retrieval, the models rank about as well as trial history, a count of
each target's past trials in the same therapeutic area, also after their
training cutoffs. Evidence dated before the
decision adds at most a small gain, before or after the models' training
cutoffs. The newest graph raises five of six models clearly above that
baseline, but its paths usually include the answer: the pair's own trial, dated
after the decision. Evaluations of target prioritisation should date both
features and labels, with evidence from before the decision and outcomes after
it.
\end{abstract}

\begin{figure}[t]
\centering
\includegraphics[width=0.50\textwidth]{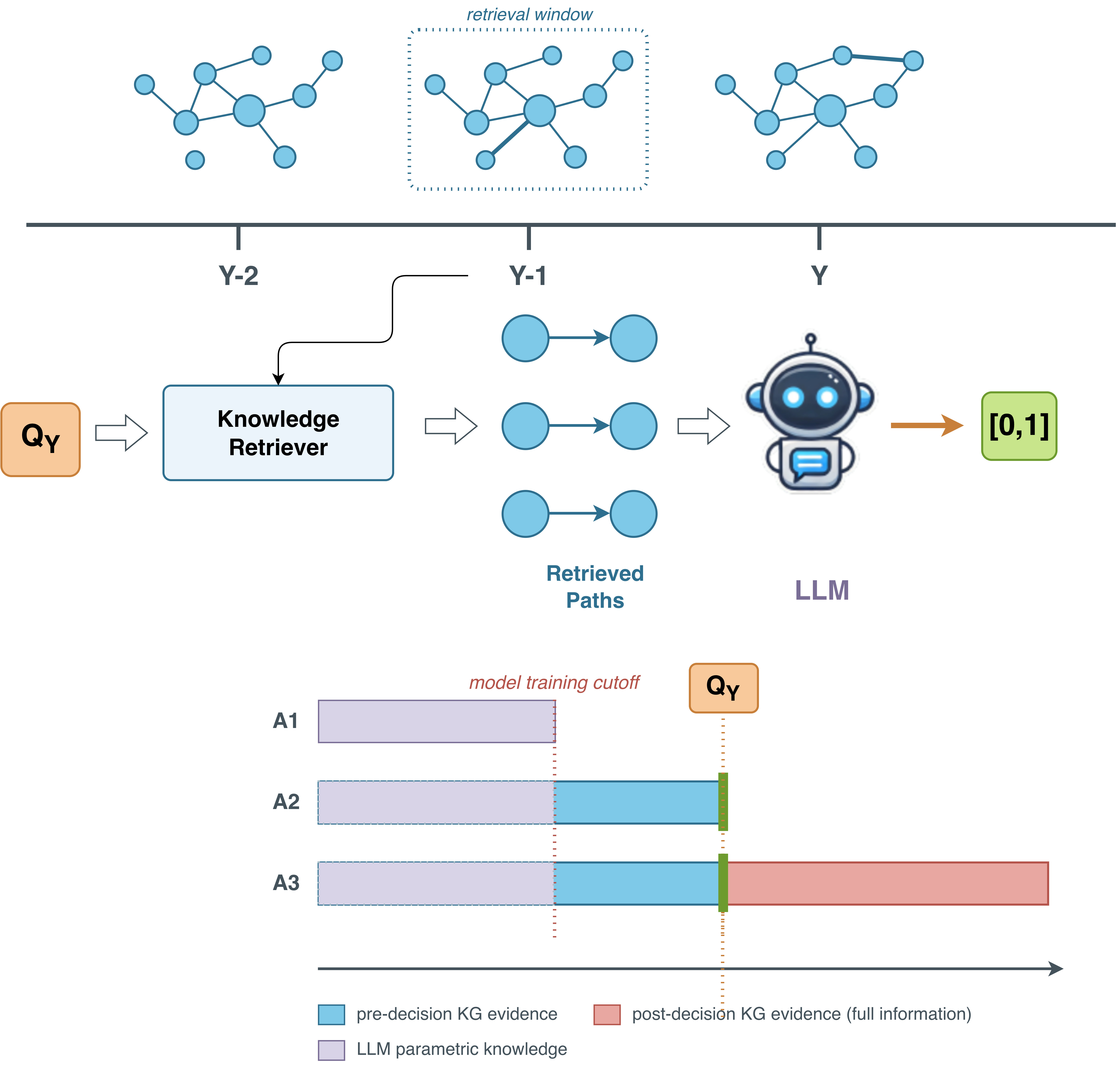}
\caption{\textbf{Overview.} For a target--disease pair queried for year $Y$,
we retrieve evidence paths from the dated knowledge graph and prompt a language
model to score the pair. \textbf{Top:} annual graph snapshots. The snapshot at $C=Y-1$ is
retrieved, excluding the predicted event. \textbf{Bottom:} the access conditions
relative to the model's training cutoff: no retrieval
with descriptions (A1), dated graph (A2) and the newest snapshot (A3).}
\label{fig:overview}
\end{figure}

\section{Introduction}

Most diseases still lack a treatment: only $612$ of the $11{,}158$ diseases in
the Human Disease Ontology have an approved drug in any region
\citep{gordillo2024disease}, and about $5\%$ of rare diseases have an
FDA-approved drug \citep{fermaglich2023}. Treating these diseases needs targets that are new to
each disease, yet
research concentrates on proteins that are already well studied
\citep{edwards2011,kustatscher2022}. For diseases with unmet need, choosing
which novel target to trial is a costly decision. Lack of efficacy is the main
cause of clinical failure, and weak preclinical evidence linking target to
disease is a major contributor \citep{hingorani2019}. Genetic support for the
link raises the success rate 2.6-fold, most in Phase~II and III, where efficacy
is tested \citep{minikel2024}. AI platforms
\citep{kamya2024} and, increasingly, large language models (LLMs) are used to
rank candidate targets for this decision.

TargetBench \citep{leung2026targetbench}
includes frontier models among the target-identification systems it
benchmarks, and \citet{bhatnagar2025} rank Alzheimer's disease targets with a
search-augmented LLM. Both validate against targets already in clinical
development, without regard to time, whereas other
target-ranking models restrict their training data or evidence to a cutoff date
so that they cannot score by recalling known links
\citep{narganes2023,narganes2024,patel2024,otrec2025}. Good scientific
reasoning does not by itself make good forecasts \citep{cusp2026}, and no one
has tested whether LLMs can reason from the evidence available at a decision to
the novel target--disease hypotheses that will next enter clinical development.
A model's training
cutoff offers a temporal grounded test, because events after it are absent
from its training data \citep{cusp2026,dai2024dailyoracle}. Retrieval of relevant evidence
complicates this test: it can raise accuracy \citep{cusp2026,dai2024dailyoracle},
and it can also bring post-cutoff evidence back in. We therefore score each
model against its own training cutoff, with and without retrieval, using a dated
graph to separate evidence available at the decision from evidence that arrived
after it.

We run this test on THBKG \citep{thbkg2026}, a temporal biomedical knowledge graph that can be
rewound to any decision point. For each target--disease pair, we retrieve
multi-hop paths from the graph as it stood at that point and ask an open-weight
LLM to score the pair (Figure~\ref{fig:overview}). A ladder of access conditions
(\S\ref{sec:method}) adds one source of information at a time, from the
model's parametric knowledge, through the dated graph, to the newest graph,
decomposing performance into the contributions of parametric knowledge,
decision-time evidence and post-decision evidence.

This paper contributes a \textbf{decision-time evaluation framework} for
novel target prioritisation, which scores each model against its own training
cutoff and bounds the label, the retrieved graph evidence and the negatives at
the decision point.\footnote{This version re-runs the evaluation in response to
review. Appendix~\ref{app:changes} lists what changed from the reviewed
version.} Applied to six open models, it shows that \textbf{decision-time
knowledge forecasts only as well as trial history}, a count of past trials: any
drop in the models' score after the training cutoff is at most $0.035$ AUROC, so most of it reflects plausibility
rather than memory of the trial, and evidence dated before the decision adds
little, before or after the cutoff. It also shows
that \textbf{gains from the newest graph are hindsight, not foresight}: the
models rise clearly above trial history, but the graph's paths usually contain
the pair's own trial, dated after the decision.

\section{Method}
\label{sec:method}

\paragraph{Data.} A query is a disease $d$ and a decision year
$Y\in\{2021,\ldots,2026\}$, with snapshot year $C=Y-1$. Positives are the targets whose
first trial for $d$ in the Open Targets \texttt{clinical\_precedence} channel
\citep[release 26.03;][]{ochoa2023opentargets} falls in $Y$ (a one-year horizon); they may have
earlier evidence of association, but no earlier trial for $d$. Negatives are
drawn per pair without future information (time-aware negatives): the
pool is every protein-coding target trialled by $C$ for a disease sharing a
therapeutic area with $d$, minus targets trialled for $d$ by $Y$, so a target first
trialled for $d$ after $Y$ remains an eligible negative. We sample ten negatives
per positive uniformly (median pool $\approx 1{,}050$), giving $1{,}373$ queries and
$6{,}394$ positives (Appendix~\ref{app:stats}); the 2026 slice covers trials
registered by May. Descriptions, therapeutic
areas and the ontology come from the undated Open Targets 26.03 release.

\paragraph{Retrieval and scoring.} We retrieve from THBKG \citep{thbkg2026}, a
temporal biomedical knowledge graph that serves as an oracle of what was known
when: its as-of query $\mathcal{G}[T]$ returns the subgraph dated before $T$,
each edge carrying a relation type and
an evidence score. For each candidate we extract the top 8 multi-hop paths ($\le 3$ hops) to the
disease with K-Paths \citep{kpaths2025}. We score
six open-weight, instruction-tuned models: Llama~3.1-8B/70B (training cutoff
December 2023) and Qwen~3-8B/32B and Gemma~2-9B/27B (June 2024;
approximate). We ask each model whether the target will enter its first trial for the
disease in $Y$, given the pair and its evidence (Appendix~\ref{app:prompts}).
We rank each query's candidates by the model's probability of answering Yes
rather than No (Eq.~\ref{eq:pyes}).

\paragraph{Conditions.} Following CUSP's base/pre-cutoff/full-information
contrast \citep{cusp2026}, we vary only what accompanies the pair
(Table~\ref{tab:conditions}), adding one source of information per step:
target and disease descriptions without retrieval (\textbf{A1});
leak-free paths from the dated graph $\mathcal{G}[C]$ (\textbf{A2}); and the
newest snapshot (\textbf{A3}), a full-information ceiling that for most positives
contains the pair's own trial edge, i.e.\ the label. In CUSP's terms, A2$-$A1 is the
knowledge gap and A3$-$A2 the forecasting gap. Each model is scored on
every year against its own training cutoff, so the pre-/post-cutoff contrast is
within-model. Every positive is novel to the \emph{evidence} by construction,
since its trial is absent from $\mathcal{G}[C]$. Post-cutoff positives are also
novel to the \emph{model}, since their first trial postdates its training data.

\paragraph{Evaluation.} We compare every condition with \textbf{trial history},
a model-free baseline that ranks each candidate by the number of diseases in
$d$'s therapeutic area that the candidate was trialled for by $C$ (other trial counts and genetic evidence
in Appendix~\ref{app:vsheur}). Metrics are per query,
averaged over queries: AUROC, average precision (AP) and MRR, with 95\% CIs from 2{,}000 query resamples; chance is
$0.5$ AUROC, about $0.22$ AP and $0.26$ MRR. Protocol, data statistics and prompts are in
Appendices~\ref{app:protocol}--\ref{app:prompts}.

\section{Results}
\label{sec:results}

\subsection{Models with decision-time knowledge forecast only as well as trial history}
\label{sec:label}

Without retrieval, the models already rank well above chance, and this
parametric floor, not zero, is what any retrieval gain must beat. The floor
matches trial history in AUROC and AP and slightly exceeds it in MRR
(Table~\ref{tab:headline}). Little of it comes from the prompt's
descriptions: with identifiers only (A0), the models score almost as well
(Table~\ref{tab:headline_full}). After the cutoff, where the positives are novel to
the model, any drop in the floor is at most $0.035$ AUROC (95\% CI), about a
seventh of its margin over chance, and this holds for Llama alone and without
the first post-cutoff year (Table~\ref{tab:headline};
Appendix~\ref{app:prepost}). Most of it therefore reflects plausibility rather
than memory of the trial, but it does not exceed the prioritisation that trial
history  provides.

\begin{table}[t]
\centering\scriptsize
\setlength{\tabcolsep}{4pt}
\begin{tabular}{@{}ll*{9}{c}@{}}
\toprule
& & \multicolumn{3}{c}{\textbf{AUROC}} & \multicolumn{3}{c}{\textbf{AP}} & \multicolumn{3}{c}{\textbf{MRR}} \\
\cmidrule(lr){3-5}\cmidrule(lr){6-8}\cmidrule(l){9-11}
\textbf{Model} & & A1 & A2 & A3 & A1 & A2 & A3 & A1 & A2 & A3 \\
\midrule
Llama 3.1-8B & pre & .732 & .752 & .851 & .437 & .468 & .580 & .484 & .530 & .641 \\
 & post & .727 & .731 & .854 & .437 & .448 & .593 & .473 & .489 & .641 \\
\cmidrule{1-11}
Llama 3.1-70B & pre & .752 & .766 & .897 & .468 & .478 & .672 & .515 & .522 & .715 \\
 & post & .741 & .749 & .912 & .459 & .460 & .725 & .482 & .483 & .748 \\
\cmidrule{1-11}
Qwen 3-8B & pre & .736 & .729 & .754 & .449 & .445 & .504 & .492 & .490 & .572 \\
 & post & .737 & .721 & .800 & .444 & .449 & .585 & .474 & .490 & .634 \\
\cmidrule{1-11}
Qwen 3-32B & pre & .769 & .784 & .919 & .498 & .513 & .706 & .548 & .561 & .757 \\
 & post & .768 & .773 & .942 & .502 & .504 & .785 & .544 & .538 & .814 \\
\cmidrule{1-11}
Gemma 2-9B & pre & .726 & .748 & .934 & .463 & .471 & .786 & .502 & .520 & .830 \\
 & post & .719 & .728 & .943 & .458 & .458 & .803 & .477 & .494 & .818 \\
\cmidrule{1-11}
Gemma 2-27B & pre & .766 & .756 & .847 & .504 & .484 & .592 & .556 & .540 & .643 \\
 & post & .749 & .743 & .851 & .481 & .494 & .626 & .515 & .541 & .656 \\
\midrule
\multicolumn{2}{@{}l}{Trial history} & \multicolumn{3}{c}{.754} & \multicolumn{3}{c}{.467} & \multicolumn{3}{c}{.474} \\
\bottomrule
\end{tabular}
\caption{Per-query performance on the test sets
(2021--2026), before (pre) and after (post) each model's training cutoff:
December 2023 for Llama ($926$ / $447$ queries) and about June 2024 for Qwen and
Gemma ($1{,}169$ / $204$; approximate). 
Chance is $0.5$ AUROC, $0.22$ AP and $0.26$ MRR. The trial-history row uses all
queries; per-split values differ by at most $0.03$. Other conditions are
in Table~\ref{tab:headline_full}. }
\label{tab:headline}
\end{table}

\begin{figure}[t]
\centering
\includegraphics[width=5.31in]{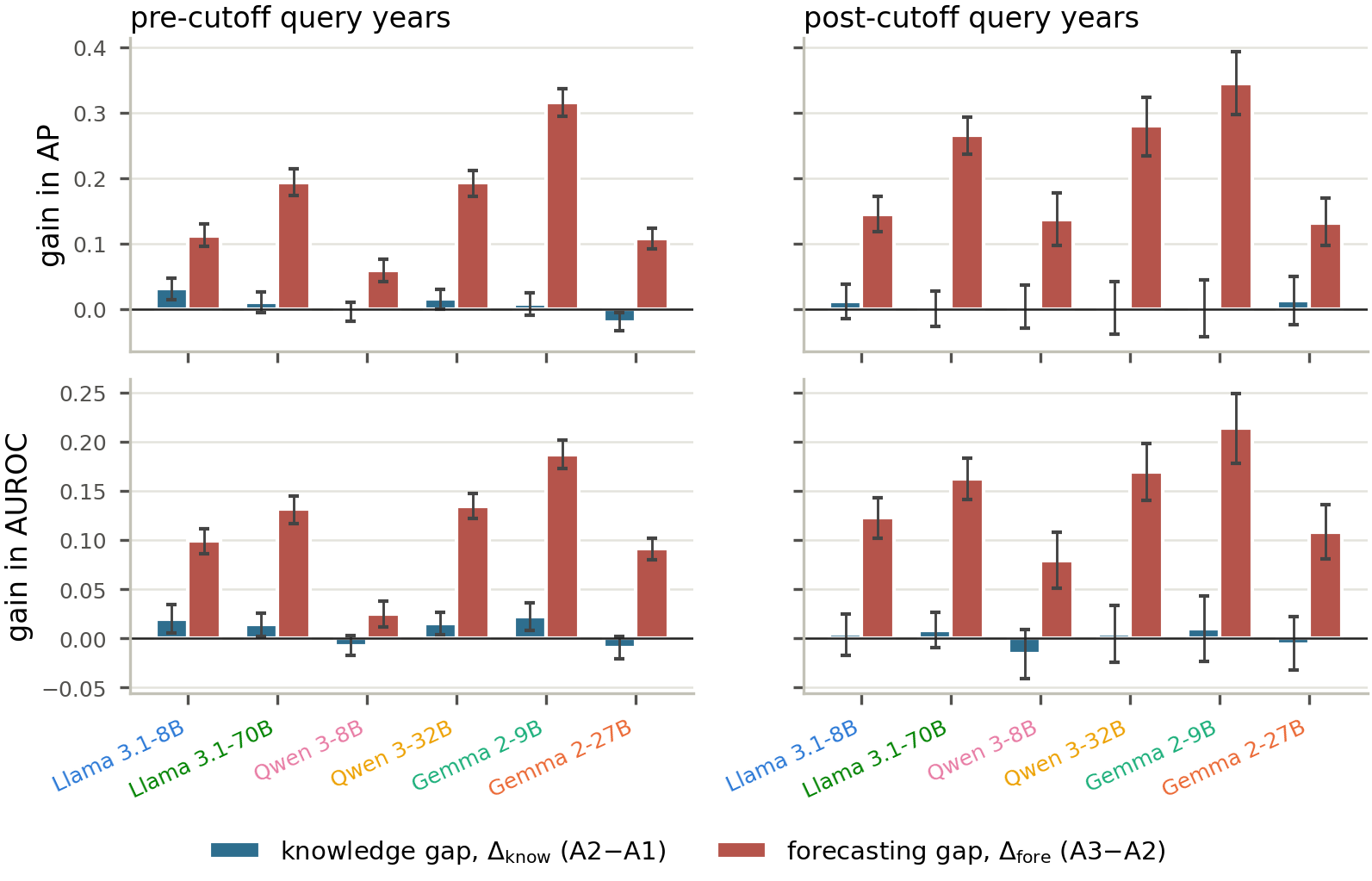}
\caption{Knowledge gap ($\Delta_\mathrm{know}$, A2$-$A1, blue) and forecasting gap
($\Delta_\mathrm{fore}$, A3$-$A2, red) per model, as defined by CUSP \citep{cusp2026},
before and after the model's training cutoff, in per-query AP (top) and AUROC
(bottom), 2021--2026 (queries matched across conditions; whiskers are 95\% CIs
from 2{,}000 query resamples). The forecasting gap is filled by post-decision
evidence, which usually includes the pair's own trial (\S\ref{sec:leak}). Model
names are coloured as in Figure~\ref{fig:decay_ap}.}
\label{fig:gaps_main}
\end{figure}

Evidence dated before the decision adds little on top of this floor
(Figure~\ref{fig:gaps_main}). Before the training cutoff, the dated graph
improves AUROC slightly for four of six models. After it, where the positives
are novel to the model and the paths are its only window onto later biology, the gain is significant for no model. The dated graph
therefore neither substitutes for missing knowledge after the cutoff nor
unlocks under-used parametric knowledge before it, the pre-cutoff surge that
CUSP reports for scientific forecasting \citep{cusp2026}.

\subsection{Gains from the newest graph are hindsight, not foresight}
\label{sec:leak}

A3 raises every model and lifts five of
six clearly above trial history (Figure~\ref{fig:gaps_main},
Table~\ref{tab:headline}), whereas without leaked evidence the models stay close
to it. The source of this gain is visible in the evidence itself: for most
positives, but almost no negatives, the eight paths shown in A3 contain a direct
trial edge between the queried target and disease, registered after the
decision. A retriever over today's graph thus often shows the model the outcome
it is asked to predict.

\begin{figure}[t]
\centering
\includegraphics[width=5.30in]{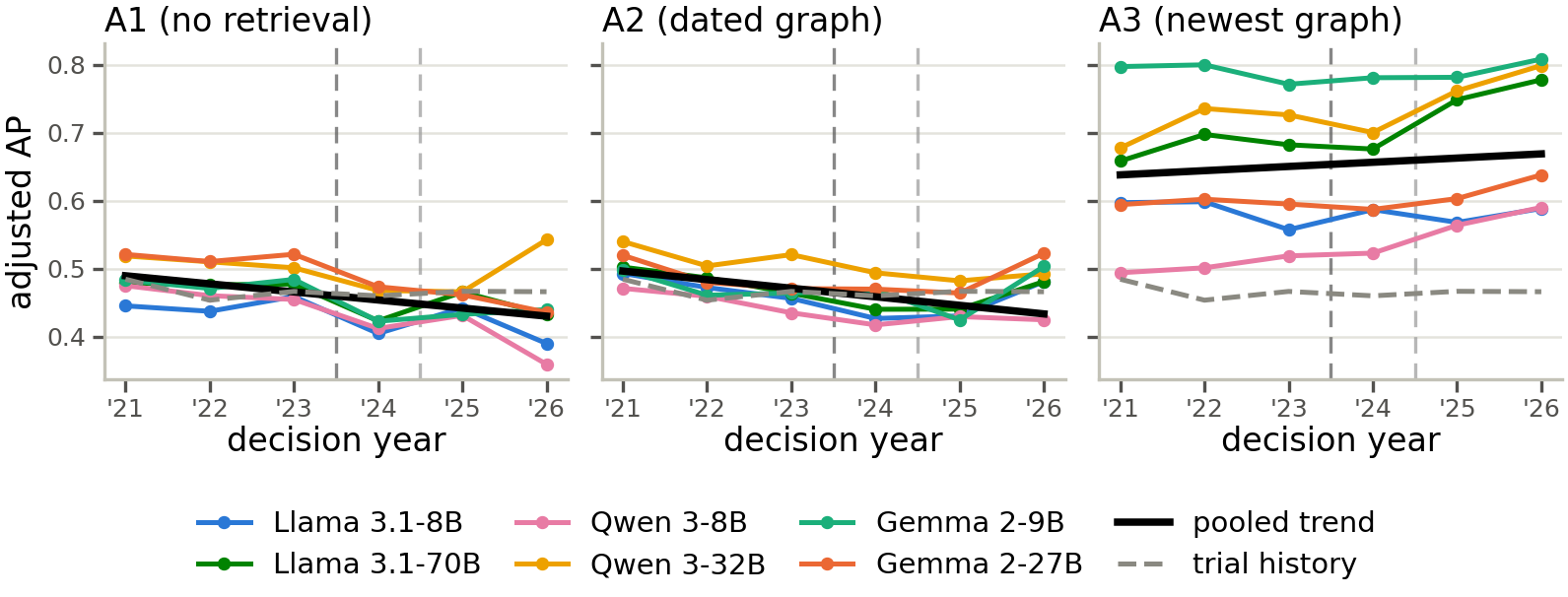}
\caption{Per-query AP by decision year,
2021--2026, adjusted for query size (residualised on log
list length within each condition), one curve per model. Black: pooled linear
trend from a query-level regression with model fixed effects (standard errors
clustered by disease). Pooled slopes in AP per year: A1 $-0.012^*$, A2
$-0.013^*$, A3 $+0.0062$; trial history $-0.0035$ ($^*$ 95\% CI excludes zero).
Grey dashed: trial history, adjusted the same way. Vertical dashed lines
mark the Llama and Qwen/Gemma cutoffs.}
\label{fig:decay_ap}
\end{figure}

Over decision years, leak-free performance declines in adjusted AP, though not significantly against trial history,
while A3 does not decline, as expected if the leaked edge is equally decisive in
every year (Figure~\ref{fig:decay_ap}; Appendix~\ref{app:expanding}).

\section{Discussion and conclusion}
\label{sec:discussion}

A target-prioritisation model can look strong and still be reading the answer.
Without retrieval, the models rank well above chance, about as well as trial
history, and from plausibility rather than memory of the trial, since their score
changes little after their training cutoffs (Appendix~\ref{app:prepost}). Dated
evidence does not turn this into foresight: the knowledge gap \citep{cusp2026} is
small. The forecasting gap is large,
but it comes from the pair's own later trial appearing in the evidence.
Benchmarks that retrieve from today's evidence are therefore at risk of scoring
hindsight as foresight.

Evaluations of target prioritisation should therefore bound the
\emph{evidence} as well as the label. Target ranking with general-purpose LLMs
has so far been validated with neither bound \citep{leung2026targetbench,bhatnagar2025}, in one
case with retrieval from the live web \citep{bhatnagar2025}, yet trial records
carry registration dates, so a date filter on the retrieved evidence removes the
leak.

These conclusions hold for open models up to 70B under this protocol. We read a
single Yes/No answer without chain-of-thought, so we test whether evidence
improves the model's judgement, not its explicit reasoning; frontier and
reasoning models remain untested. Before the cutoff, plausibility and residual memory of
a trial cannot be separated pair by pair, although the floor's stability across
the cutoff limits how much memory can contribute. We will next carry the same
cutoff discipline from evaluation into training, finetuning only on events
before each decision point.

\FloatBarrier
\acksection
This work was supported by the UKRI Biotechnology and Biological Sciences Research
Council [BB/X511833/1,BB/Y513416/1], the Digital Environment Research Institute (DERI), Queen Mary
University of London, and Recursion Pharmaceuticals Inc. We thank
Apocrita~\citep{king2017apocrita} and its ITS team for providing and maintaining the
HPC resources.

\textbf{Competing interests.} Mani Mudaliar is an employee of Recursion
Pharmaceuticals Inc. The other authors declare no competing interests.

\bibliographystyle{plainnat}
\bibliography{refs}

\appendix

\section{Changes from the reviewed version}
\label{app:changes}

Following review, we re-ran every condition on a corrected benchmark. The
reviewed version drew negatives with a rule that excluded any target entering a
later trial for any disease; this used post-decision information and left easier
negatives in earlier years. All results now use per-pair, time-aware negatives
at $1\!:\!10$ over 2021--2026 (\S\ref{sec:method}). The prompt now asks about the decision year itself rather
than ``after $C$''. Inference uses Hugging Face transformers in place of
llama.cpp Q4\_K\_M. Average precision replaces precision@$5$, whose ceiling
depends on the number of positives.

We added analyses requested in review or needed to read them: a trial-history
baseline (\S\ref{sec:label}); an identifiers-only condition (A0) bounding the effect of
the undated descriptions; and a check of whether A3's paths contain the queried
pair's own trial edge.

The main finding is unchanged: the dated graph adds little, and the
newest graph's advantage rests on post-decision trial records that show the
label. One finding changed in substance. The reviewed version reported that leak-free performance
declines over time; under its sampler a model-free ranking declined as well, so that evidence could not separate the models from
the benchmark. With time-aware negatives the models' adjusted AP still declines, but
trial history drifts slightly too, and the decline relative to it is not
significant (Appendix~\ref{app:expanding}).

\section{Protocol detail}
\label{app:protocol}

\paragraph{Cutoff axis.} A year that is post-cutoff for one model is
pre-cutoff for another, and no single boundary makes every model ignorant, hence
the within-model reading. Open weights remove the confound of silent
proprietary updates, but four of six cutoffs are approximate. We treat release
dates as conservative upper bounds.

\paragraph{Access conditions.} All conditions share the pointwise prompt and
scoring and differ in the description text, in the retrieval graph and which
edges it may contain, and in the stated evidence date (Table~\ref{tab:conditions};
Appendix~\ref{app:prompts}). Paths are at most three hops throughout. Each path
edge carries an evidence-score annotation (e.g.\ \texttt{--relation[score=0.84]-->});
A2 strips it, leaving relation labels and topology, while A3 shows it.
A3$-$A2 therefore combines the newest graph with the visible scores.
A0 uses the A1
prompt with both description fields replaced by ``(no description
available)''.

\begin{table}[H]
\centering\small
\begin{tabular}{@{}llll@{}}
\toprule
\textbf{Cond.} & \textbf{Retrieval graph} & \textbf{Post-decision edges} & \textbf{Role} \\
\midrule
A0       & none (identifiers only)  & ---            & no-text reference \\
A1       & none (descriptions)      & ---            & no-retrieval reference \\
A2       & snapshot $\mathcal{G}[C]$ & none       & leak-free gain \\
A3       & newest snapshot          & all            & full information, incl.\ label \\
\bottomrule
\end{tabular}
\caption{The access conditions. They differ in the description text, the
retrieval graph, the stated evidence date (Appendix~\ref{app:prompts}) and, for
A3, the edge-score annotations. Prompt, scoring, pairs, and the $\le 3$-hop path
budget are held fixed.}
\label{tab:conditions}
\end{table}

\paragraph{Metrics.} Uniform negatives make AUROC unbiased for the whole pool.
AP and MRR depend on the $1\!:\!10$ ratio. Chance is $0.261$ MRR and $0.219$ AP.

\section{Dataset and retrieval statistics}
\label{app:stats}

\begin{figure}[H]
\centering
\begin{minipage}[t]{0.49\textwidth}\centering
  \textbf{(a)}\\
  \includegraphics[width=2.50in]{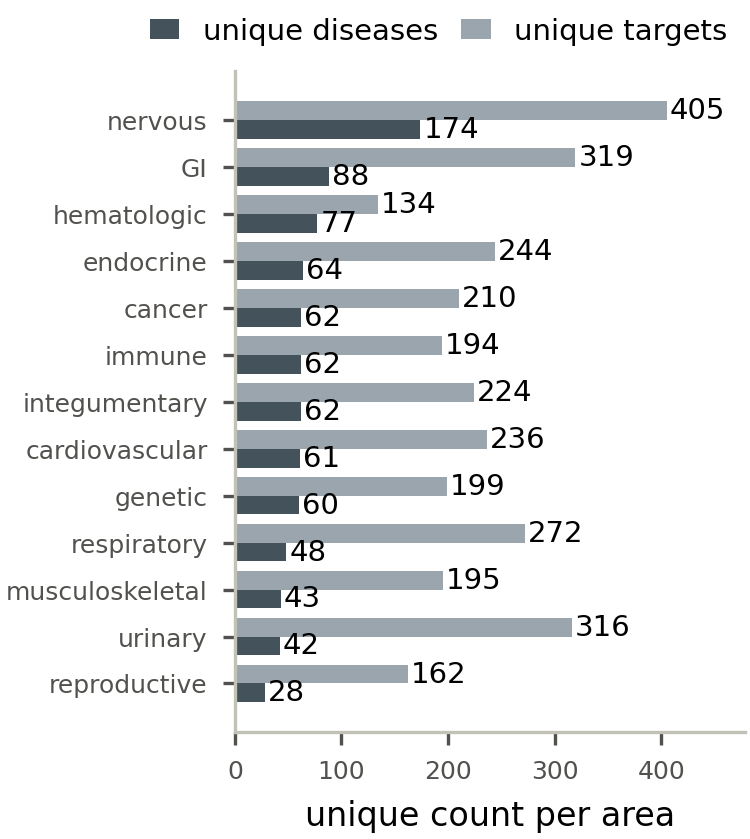}
\end{minipage}\hfill
\begin{minipage}[t]{0.49\textwidth}\centering
  \textbf{(b)}\\
  \includegraphics[width=2.50in]{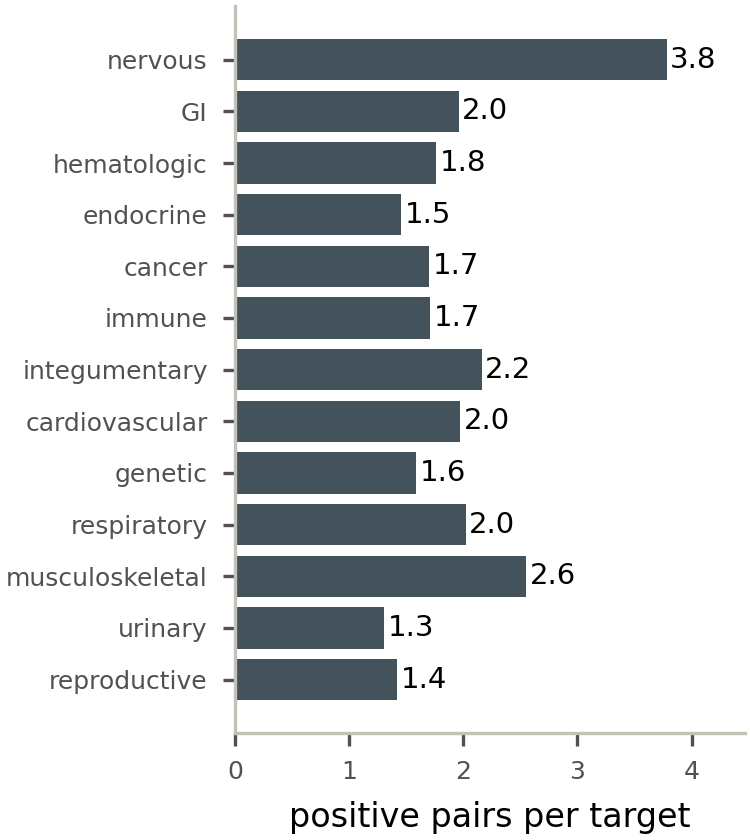}
\end{minipage}\\[6pt]
\textbf{(c)}\\
\includegraphics[width=4.75in]{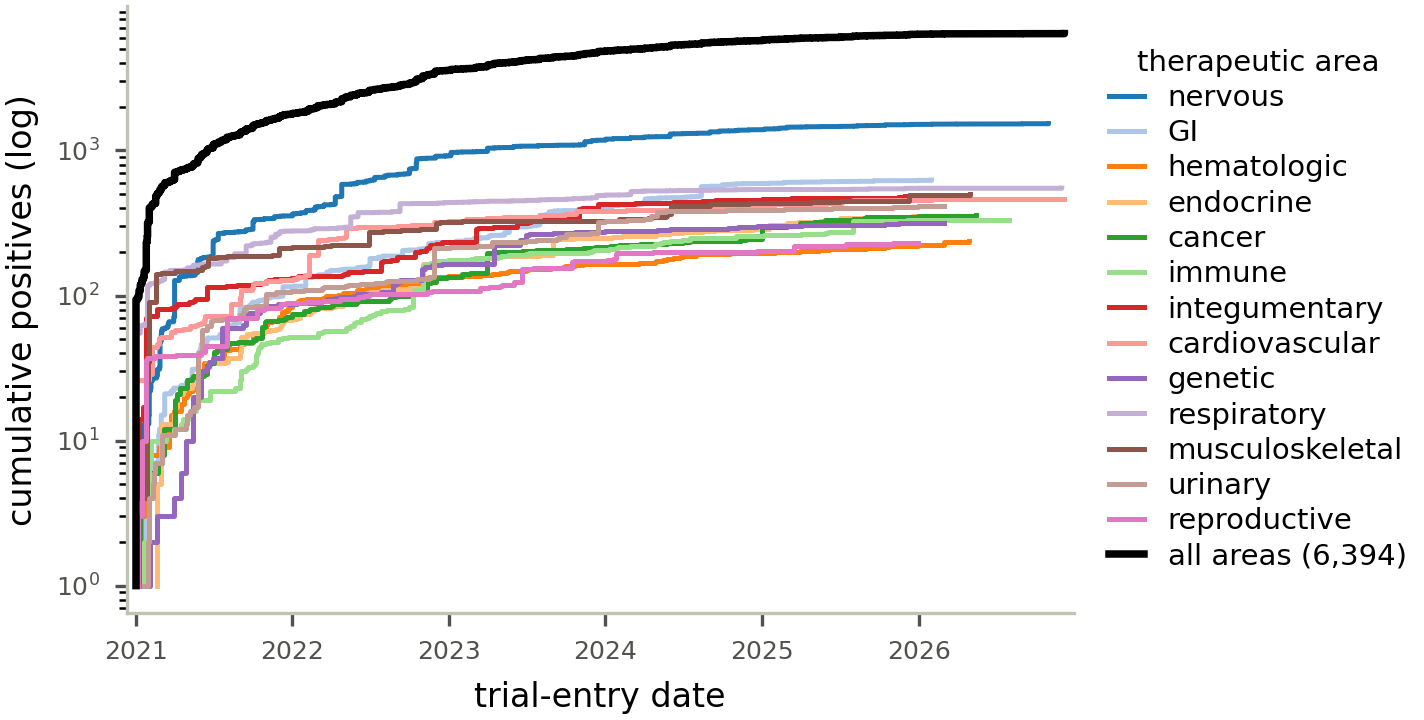}
\caption{Composition of the scored positive set. \textbf{(a)} unique diseases and
unique targets per therapeutic area (values annotated). \textbf{(b)} target reuse
(positive pairs per unique target). \textbf{(c)} cumulative
positives (first trials for the disease) per area by exact trial-entry date (2021--2026, log scale). The bold
line is all areas ($6{,}394$). Nervous-system disease dominates, and every area's
curve flattens toward the present as recent first trials are
progressively censored.}
\label{fig:dataset}
\end{figure}

The six yearly test sets hold $357$, $317$, $252$, $243$, $171$ and $33$
queries and $1{,}785$, $1{,}781$, $1{,}269$, $885$, $613$ and $61$ positives
for 2021 through 2026, with ten negatives per positive ($70{,}334$ scored pairs);
2026 covers trials registered by May. The median query has two positives in
2021--2025, but the mean falls over time ($5.0$ in 2021, $3.6$ in 2024--25),
which is why the temporal-decay estimates in
Appendix~\ref{app:expanding} adjust for query size rather than reading raw per-year
levels. Figure~\ref{fig:dataset} shows the positives by therapeutic area and over
time.

\section{Prompt templates}
\label{app:prompts}

All conditions share one pointwise template. The only differences are the
system line and the presence of the evidence-paths block. Curly-brace fields
are filled per pair. The A1 system line is:

\begin{quote}\small\ttfamily
You are forecasting whether a drug target will enter a clinical trial as a
therapeutic hypothesis for a specific disease. You have no access to
information beyond \{cutoff date\}.
\end{quote}

\noindent
and the A2/A3 system line replaces the second sentence with ``All evidence
presented below was available as of \{cutoff date\}. You have no access to
information beyond this date.'' The user message is:

\begin{quote}\small\ttfamily
TARGET: \{symbol\} (\{Ensembl id\})\\
\hspace*{1em}Protein class: \{class\}\\
\hspace*{1em}Known functions: \{description\}\\[2pt]
DISEASE: \{name\} (\{ontology id\})\\
\hspace*{1em}Therapeutic area: \{area\}\\
\hspace*{1em}Known biology: \{description\}\\[2pt]
EVIDENCE PATHS connecting \{symbol\} to \{name\}:\hfill\textrm{\emph{(A2/A3 only)}}\\
Path 1: \{gene\} (target) --associated\_with--> \{node\} (pathway)
--involved\_in--> \{disease\} (disease)\\
\ldots\\[2pt]
Based on the evidence above (available as of \{graph cutoff\}), will
\{symbol\} enter its first clinical trial as a therapeutic target for \{name\}
in \{decision year\}?\\[2pt]
Answer with a single word: Yes or No. Output only that word, with no
punctuation or explanation.
\end{quote}

\noindent
The question asks about first entry \emph{in} the decision year $Y$, matching the
label. The reviewed version asked about entry ``after'' the snapshot year. In A0
and A1 the paths block is absent, the question reads ``Based on your knowledge
available as of December 2023\ldots'', and the system line's \{cutoff date\} is
also December 2023, for every model and year. A0 additionally shows
``(no description available)'' in both description fields. The stated evidence date is
December of $C$ for A2 and December 2026 for A3. Path edges carry bracketed
evidence-score annotations (e.g.\ \texttt{--relation[score=0.84]-->}); the A2
serialisation strips them, leaving relation labels and topology alone, and A3
keeps them. Scoring
reads the first generated token's log-probabilities as
\begin{equation}
  p_\text{yes} = \frac{\exp(\mathrm{lse}_\text{yes})}
                      {\exp(\mathrm{lse}_\text{yes}) + \exp(\mathrm{lse}_\text{no})},
  \label{eq:pyes}
\end{equation}
\noindent
where $\mathrm{lse}_\text{yes}$/$\mathrm{lse}_\text{no}$ are the log-sum-exp of
the ``Yes''/``No'' surface-form token variants.

\section{Full results}
\label{app:prepost}

Table~\ref{tab:headline_full} repeats Table~\ref{tab:headline} for every
condition, including identifiers only (A0).

\paragraph{The floor across the cutoff.} Without retrieval (A1), AUROC after
each model's training cutoff differs from before it by $-0.007$ (95\% CI
$-0.035$ to $+0.021$; pooled over models with model fixed effects, standard
errors clustered by disease), and relative to trial history on the same queries
by $+0.016$ ($-0.024$ to $+0.056$). No model's floor changes significantly.
Post-cutoff positives are novel to the model, so a floor that depended on
memory of the trial would drop after the cutoff. Two choices could hide a drop:
approximate cutoffs, and trials announced before the cutoff but started after
it. For Llama alone, whose cutoff is documented, the change is $-0.008$
($-0.035$ to $+0.019$). Leaving out each model's first post-cutoff year, where
pre-announced trials are most likely, it is $+0.012$ ($-0.027$ to $+0.050$), and
$+0.047$ ($-0.012$ to $+0.106$) relative to trial history.

\begin{table}[H]
\centering\scriptsize
\setlength{\tabcolsep}{3pt}
\begin{tabular}{@{}ll*{12}{c}@{}}
\toprule
& & \multicolumn{4}{c}{\textbf{AUROC}} & \multicolumn{4}{c}{\textbf{AP}} & \multicolumn{4}{c}{\textbf{MRR}} \\
\cmidrule(lr){3-6}\cmidrule(lr){7-10}\cmidrule(l){11-14}
\textbf{Model} & & A0 & A1 & A2 & A3 & A0 & A1 & A2 & A3 & A0 & A1 & A2 & A3 \\
\midrule
Llama 3.1-8B & pre & .714 & .732 & .752 & .851 & .417 & .437 & .468 & .580 & .468 & .484 & .530 & .641 \\
 & post & .709 & .727 & .731 & .854 & .432 & .437 & .448 & .593 & .470 & .473 & .489 & .641 \\
\cmidrule{1-14}
Llama 3.1-70B & pre & .733 & .752 & .766 & .897 & .448 & .468 & .478 & .672 & .489 & .515 & .522 & .715 \\
 & post & .727 & .741 & .749 & .912 & .456 & .459 & .460 & .725 & .482 & .482 & .483 & .748 \\
\cmidrule{1-14}
Qwen 3-8B & pre & .730 & .736 & .729 & .754 & .443 & .449 & .445 & .504 & .479 & .492 & .490 & .572 \\
 & post & .732 & .737 & .721 & .800 & .456 & .444 & .449 & .585 & .473 & .474 & .490 & .634 \\
\cmidrule{1-14}
Qwen 3-32B & pre & .745 & .769 & .784 & .919 & .468 & .498 & .513 & .706 & .515 & .548 & .561 & .757 \\
 & post & .766 & .768 & .773 & .942 & .510 & .502 & .504 & .785 & .557 & .544 & .538 & .814 \\
\cmidrule{1-14}
Gemma 2-9B & pre & .717 & .726 & .748 & .934 & .436 & .463 & .471 & .786 & .473 & .502 & .520 & .830 \\
 & post & .712 & .719 & .728 & .943 & .443 & .458 & .458 & .803 & .467 & .477 & .494 & .818 \\
\cmidrule{1-14}
Gemma 2-27B & pre & .741 & .766 & .756 & .847 & .466 & .504 & .484 & .592 & .512 & .556 & .540 & .643 \\
 & post & .743 & .749 & .743 & .851 & .482 & .481 & .494 & .626 & .506 & .515 & .541 & .656 \\
\bottomrule
\end{tabular}
\caption{Per-query performance under every condition, before (pre) and after
(post) each model's training cutoff; layout and queries as in
Table~\ref{tab:headline}.}
\label{tab:headline_full}
\end{table}

\subsection{Comparison with trial history}
\label{app:vsheur}

Table~\ref{tab:single} scores single label-blind baselines at $C$ on the same
queries. Trial history, the count of trials for diseases in $d$'s area, is the
strongest of four trial counts. Dated genetic evidence ranks at chance: by $C$,
only $0.9\%$ of positives have a genetic association with $d$, and $2.3\%$
with $d$ or one of its descendants. Most candidates therefore tie at zero, and
because ties are broken against the positives, its AP and MRR fall below
chance. In well-studied indications, genetic support rises along the pipeline,
from $2.0\%$ of drug mechanisms at the preclinical stage to $8.2\%$ of approved
ones \citep{nelson2015}, so it tracks later success more than first entry.

\begin{table}[H]
\centering\scriptsize
\begin{tabular}{@{}lcccc@{}}
\toprule
\textbf{Baseline (counts at $C$)} & AUROC & AP & MRR & \textbf{Positives scored}\\
\midrule
Trials for diseases in $d$'s area (trial history) & $0.754$ & $0.467$ & $0.474$ & $98.0\%$ \\
Trials for any disease & $0.702$ & $0.388$ & $0.377$ & $99.9\%$ \\
Recent trial pairs ($C-2$ to $C$) & $0.748$ & $0.405$ & $0.393$ & $97.1\%$ \\
Trials for ontology relatives of $d$ & $0.705$ & $0.368$ & $0.372$ & $57.0\%$ \\
\addlinespace[2pt]
Genetic association with $d$ & $0.507$ & $0.091$ & $0.087$ & $0.9\%$ \\
Genetic association with $d$ or a descendant & $0.513$ & $0.108$ & $0.112$ & $2.3\%$ \\
\bottomrule
\end{tabular}
\caption{Single label-blind baselines, per query, all years. Genetic
association is the Open Targets evidence score in the snapshot at $C$. The
last column is the share of positives with a nonzero score. Chance is $0.5$
AUROC, $0.22$ AP and $0.26$ MRR.}
\label{tab:single}
\end{table}

\section{Temporal decline}
\label{app:expanding}

\begin{figure}[H]
\centering
\raisebox{1.95in}{\textbf{(a)}}\,\includegraphics[width=4.48in]{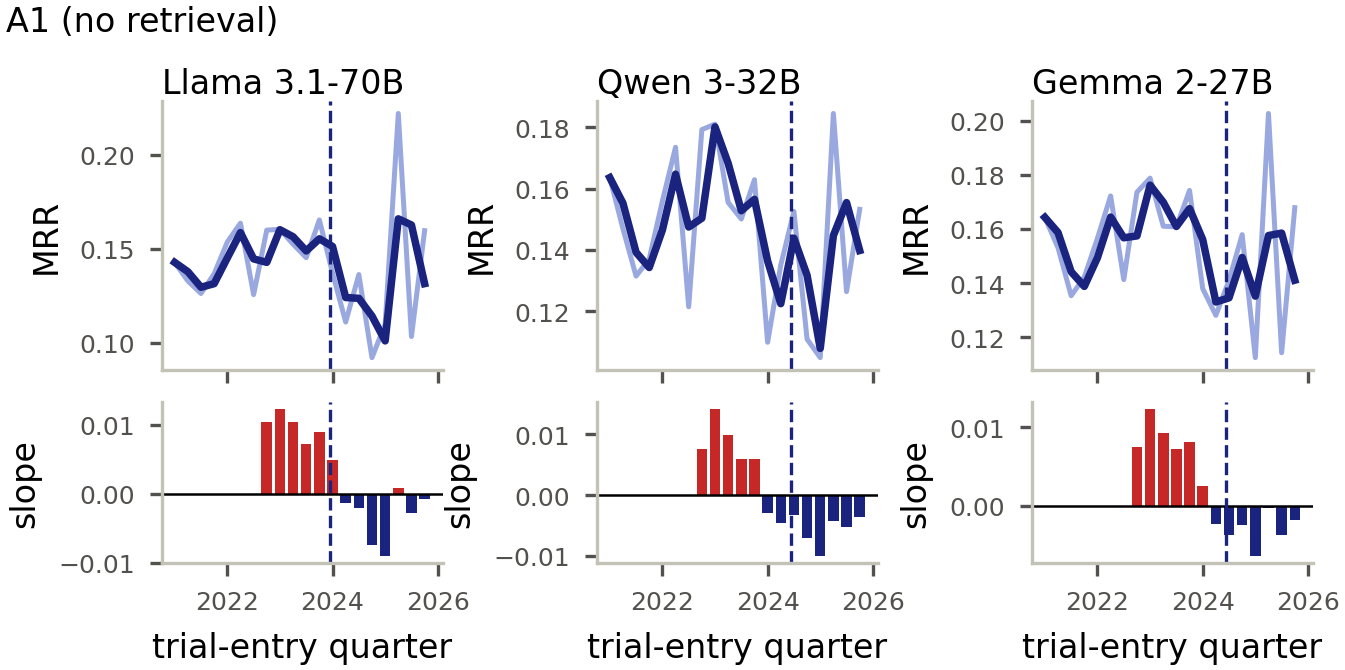}\\[2pt]
\raisebox{1.95in}{\textbf{(b)}}\,\includegraphics[width=4.48in]{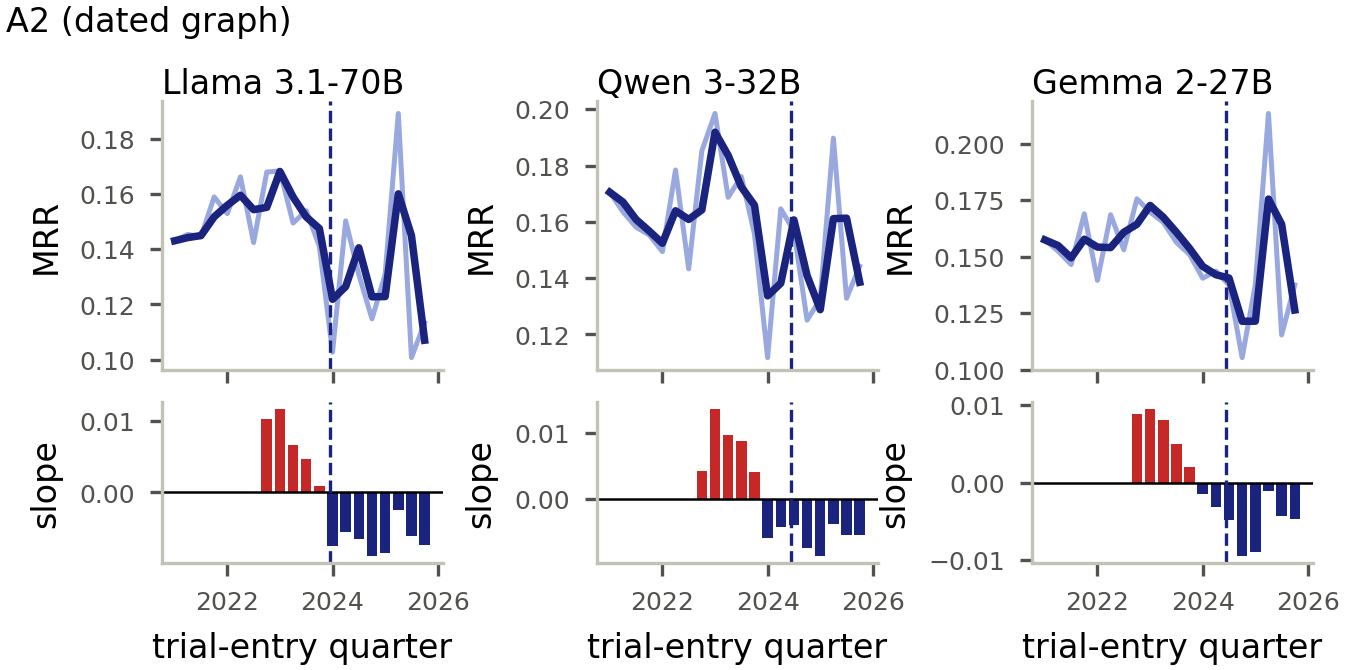}\\[2pt]
\raisebox{1.95in}{\textbf{(c)}}\,\includegraphics[width=4.48in]{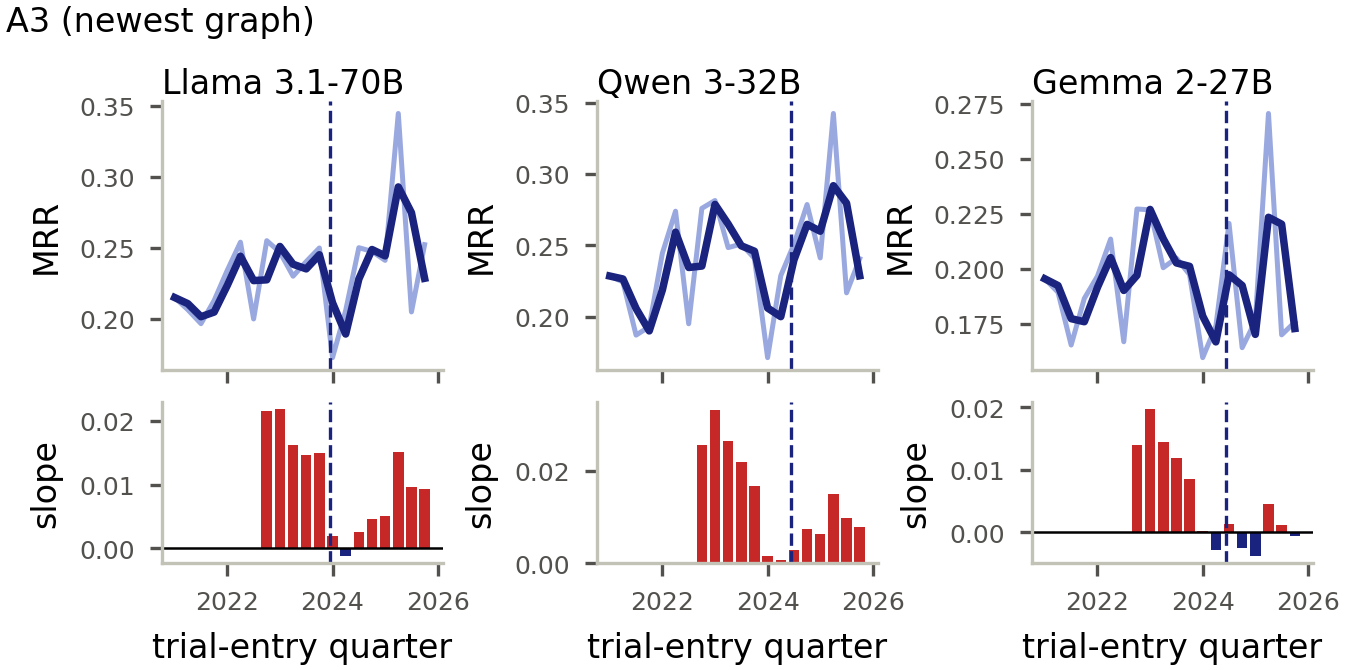}
\caption{Expanding-window regression slope of the decline for the three larger
models (70B/32B/27B): (a) A1, (b) A2 and (c) A3 (cf.\ \citealp{dai2024dailyoracle}).
Each bar is the slope of a regression on the quarterly mean of per-positive
reciprocal rank (adjusted for list length) over the quarters up to that point: first eight quarters, then
one more at a time, earlier quarters down-weighted with a half-life of six
quarters. A1 and A2 turn negative during 2024, for Qwen~3-32B before its cutoff,
whereas A3 stays positive or near zero.}
\label{fig:expanding}
\end{figure}

Over decision years, AUROC shows no significant trend for any condition or for
trial history (Table~\ref{tab:slopes}). AP depends on how many positives a query
has, and queries shrink over time, so we adjust it for query size. Adjusted AP
falls by about $0.01$ per year for A1 and A2 ($p\le0.02$) but not for A3
(Figure~\ref{fig:decay_ap}), and the decline relative to trial history is not
significant, so it cannot be attributed to the models. An expanding-window
regression \citep{dai2024dailyoracle} on per-positive reciprocal rank by trial
quarter (Figure~\ref{fig:expanding}) shows the slope turning negative during
2024 for A1 and A2, for Qwen~3-32B about two quarters before its cutoff, while
A3 stays positive or near zero.

\begin{table}[H]
\centering\scriptsize
\begin{tabular}{@{}lc@{}}
\toprule
\textbf{Series} & \textbf{AUROC per year} \\
\midrule
Trial history & $-0.0061$ (0.237) \\
LLMs, A0 & $+0.0021$ (0.615) \\
LLMs, A1 & $-0.0035$ (0.402) \\
LLMs, A2 & $-0.0052$ (0.189) \\
LLMs, A3 & $+0.0046$ (0.057) \\
\bottomrule
\end{tabular}
\caption{Yearly slope of per-query AUROC, 2021--2026, pooled over the six models
with model fixed effects. Standard errors are clustered by disease; $p$-values in
parentheses.}
\label{tab:slopes}
\end{table}

\section{Per-therapeutic-area performance}
\label{app:perta}

\begin{figure}[H]
\centering
\includegraphics[width=5.31in]{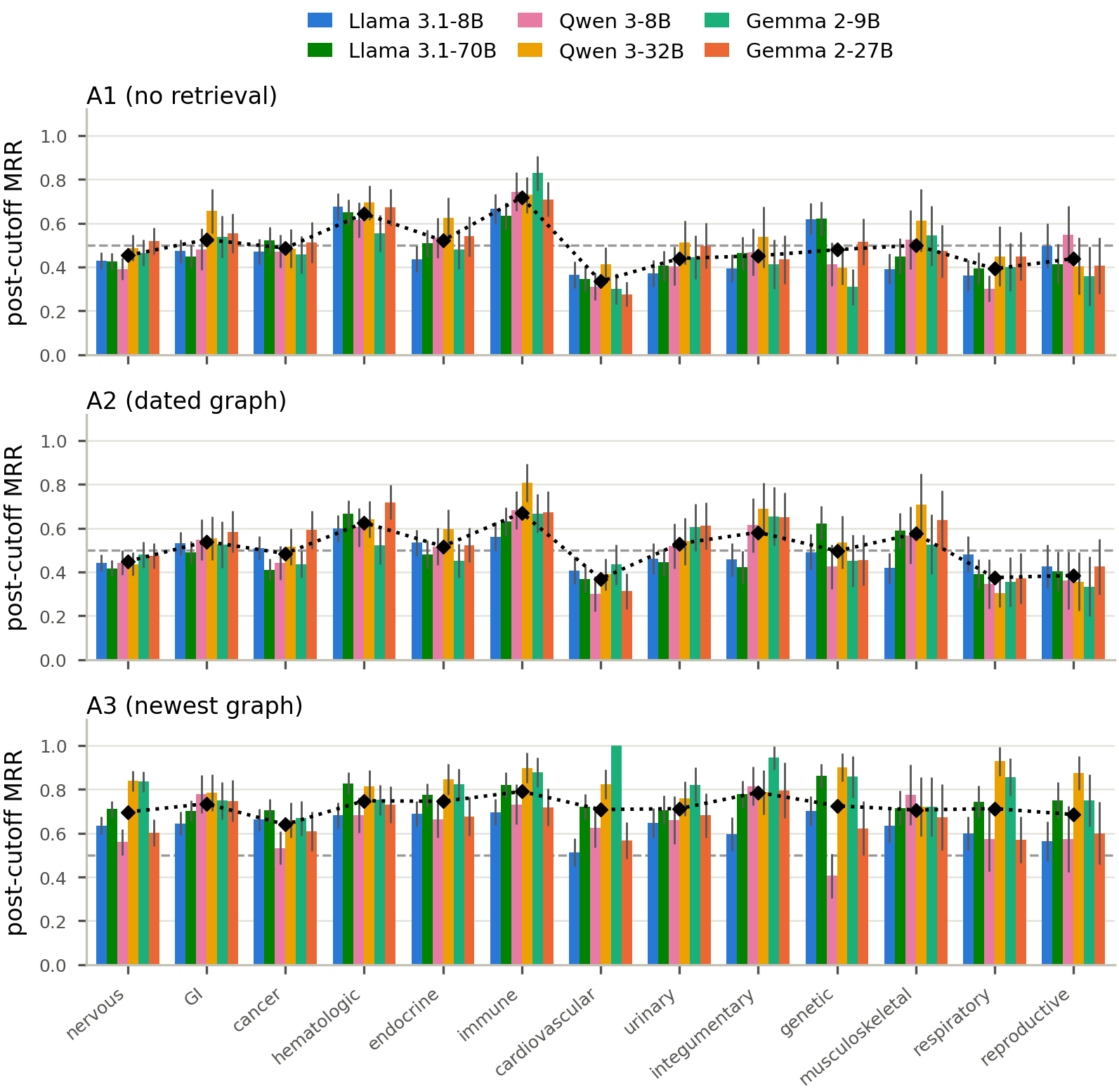}
\caption{Post-cutoff per-query MRR by therapeutic area, one panel per condition:
\textbf{(left)} A1 (no retrieval), \textbf{(middle)} A2 (dated graph),
\textbf{(right)} A3 (newest graph). Coloured bars are the six individual models (error bars SEM
over queries); the black diamond is the cross-model mean and the dotted line
connects the means across areas. The dashed grey line marks MRR $0.5$ for
reference (chance is $\approx0.26$). Areas are ordered by
post-cutoff support, shared across panels.}
\label{fig:perta_prepost}
\end{figure}

Figure~\ref{fig:perta_prepost} splits the post-cutoff per-query MRR by
therapeutic area and model for the three conditions across the thirteen canonical areas
(disease-to-area labels resolved from the Open Targets disease ontology). Under
the leak-free conditions (A1/A2) the cross-model mean is highest for immune and
haematologic diseases ($0.63$--$0.72$) and lowest for cardiovascular and
respiratory diseases ($0.34$--$0.39$), and the area ordering is broadly shared
across models. A3 is higher ($0.64$--$0.79$) across all thirteen areas.

\section{Binary calibration}
\label{app:calibration}

\begin{figure}[H]
\centering
\includegraphics[width=5.31in]{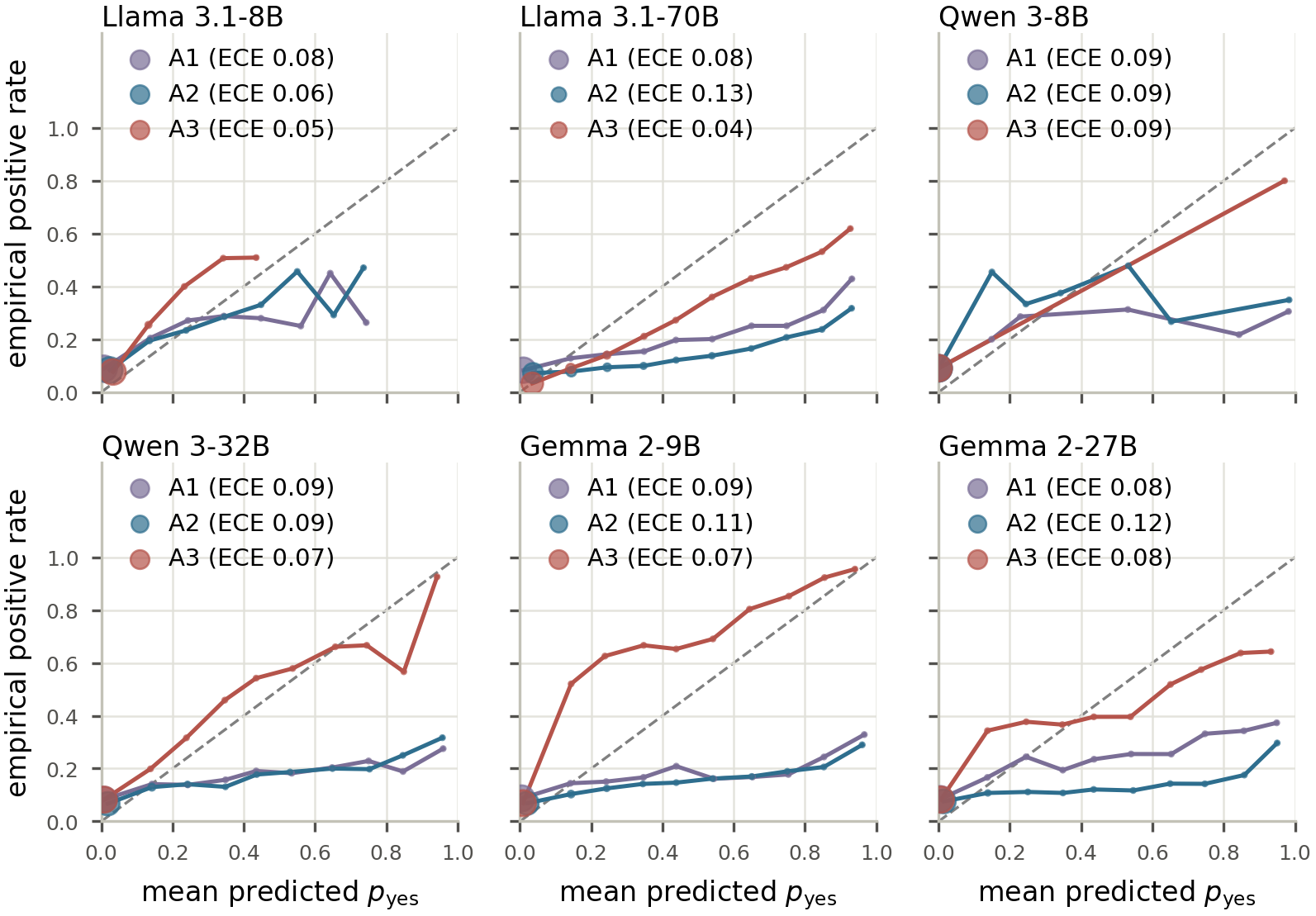}
\caption{Reliability curves for the binary $p_\text{yes}$ probe, one panel per
model, three access conditions overlaid (A1, A2, A3). Mean predicted
$p_\text{yes}$ (x) versus empirical positive rate (y), equal-width bins pooled
over all years; marker area is proportional to bin count and the dashed line is
perfect calibration. Per-condition ECE is in the legend. Bins with fewer than
15 candidates are omitted from the curve but retained in the ECE.}
\label{fig:ece}
\end{figure}

Figure~\ref{fig:ece} reports the calibration of the single-token $p_\text{yes}$
probe, per model and access condition, pooled over 2021--2026 (candidate-level,
no new inference). Binary ECE ranges from $0.04$ to $0.13$. The dated graph
worsens calibration for Llama~3.1-70B and both Gemma models (for example
Llama~3.1-70B from $0.08$ at A1 to $0.13$ at A2) and improves it for
Llama~3.1-8B, while A3 is the best calibrated condition for all six models.
Qwen~3-8B assigns almost no probability to ``Yes'' in any condition, so its ECE
equals the positive rate ($0.09$).

\section{Reproducibility}
\label{app:repro}

All inference is local with Hugging Face transformers on single A100 80GB GPUs:
bf16 weights for models up to 9B and 4-bit NF4 (bitsandbytes) for the 27B--70B
models, greedy decoding, and the answer token's log-probabilities read for the
scoring readout of Eq.~\ref{eq:pyes}. Scoring is deterministic, so repeated
runs give identical scores. The graph is reconstructed
from Open Targets release 26.03 (evidence-dated layer; Open Targets data are
CC0), and trial-entry labels derive from the release's date-resolved
clinical-precedence records, which carry the underlying ClinicalTrials.gov
registrations (public-domain registry data); no patient-level data are used
anywhere. Candidate sets are drawn at $1\!:\!10$ with a fixed seed. Bootstrap
CIs use 2{,}000 resamples of queries. As a precision check, scoring with
llama.cpp Q4\_K\_M instead changes AUROC by at most $0.019$, with no consistent
direction. Code, candidate sets and all model outputs are at
\url{https://github.com/jackysiupuichung/forecast-or-know}; they regenerate every
table and figure. The released scripts rebuild the K-path caches from THBKG
\citep{thbkg2026}.
\end{document}